# Class-Difficulty Based Methods for Long-Tailed Visual Recognition

Saptarshi Sinha[1] · Hiroki Ohashi[1] · Katsuyuki Nakamura[2]

## Abstract

Long-tailed datasets are very frequently encountered in real-world use cases where few classes or categories (known as majority or head classes) have higher number of data samples compared to the other classes (known as minority or tail classes). Training deep neural networks on such datasets gives results biased towards the head classes. So far, researchers have come up with multiple weighted loss and data re-sampling techniques in efforts to reduce the bias. However, most of such techniques assume that the tail classes are always the most difficult classes to learn and therefore need more weightage or attention. Here, we argue that the assumption might not always hold true. Therefore, we propose a novel approach to dynamically measure the instantaneous difficulty of each class during the training phase of the model. Further, we use the difficulty measures of each class to design a novel weighted loss technique called 'class-wise difficulty based weighted (CDB-W) loss' and a novel data sampling technique called 'class-wise difficulty based sampling (CDB-S)'. To verify the wide-scale usability of our CDB methods, we conducted extensive experiments on multiple tasks such as image classification, object detection, instance segmentation and video-action classification. Results verified that CDB-W loss and CDB-S could achieve state-of-the-art results on many class-imbalanced datasets such as ImageNet-LT, LVIS and EGTEA, that resemble real-world use cases.

## 1 Introduction

The rise of deep neural networks (DNN) has led to a phenomenal progress in the field of computer vision in recent years. There are many factors such as strong computational resources and large vision datasets, that are responsible for this immense success of DNN. Availability of the large-scale public datasets such as ImageNet (Deng et al., 2009), LVIS (Gupta et al., 2019) and Kinetics (Kay et al., 2017) is one of the most important factors. DNN models trained on these

Communicated by Frederic Jurie.

✉ Saptarshi Sinha
  saptarshi.sinha.hx@hitachi.com

  Hiroki Ohashi
  hiroki.ohashi.uo@hitachi.com

  Katsuyuki Nakamura
  katsuyuki.nakamura.xv@hitachi.com

[1] Intelligent Vision Research Department, Hitachi Ltd., Kokubunji 185-8601, Tokyo, Japan

[2] R&D Group, Hitachi Ltd., Kokubunji 185-8601, Tokyo, Japan

public datasets are deployed to the real-world use-cases either by fine-tuning or transfer-learning. However, the real-world data are different from these public datasets in more than one way, which causes performance drop of DNN in the real world. One important difference lies in the fact that the real-world datasets have a long-tailed (or class-imbalanced) distribution of categories while the public datasets have a more or less uniform distribution. In long-tailed datasets, few categories (called majority or head categories/classes) comprise larger number of training instances than the other categories (called minority or tail categories/classes). Training or fine-tuning a DNN model on such skewed datasets results in inference predictions biased towards the head classes. This shortcoming has stirred up a research field popularly known as 'long-tailed recognition'.

Most prior methods try to solve the issue of long-tail by making the DNN model pay more attention to the tail classes during training. The most popular ways of achieving that can be majorly classified as either sampling methods or cost-sensitive learning techniques. In a training epoch, sampling techniques such as (Chawla et al., 2002; Barua et al., 2014) sample instances from tail classes with higher frequency compared to the head classes. Cost-sensitive learning approaches such as (Cui et al., 2019; Tan et al., 2020; Cao et al., 2019) penalize the DNN model higher for making mistakes on tail class instances by assigning higher weights to the loss incurred on those samples during back-propagation. The idea of giving more focus to the tail classes seems intuitive as the general assumption is that the tail classes are always the most difficult classes due to under-representation.

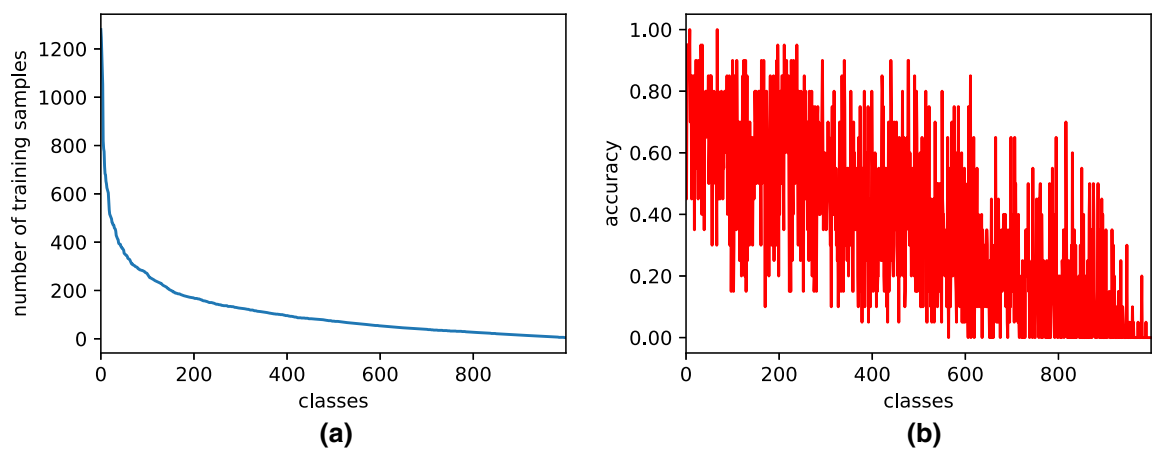

**Fig. 1** **a** Long-tailed distribution of ImageNet-LT dataset **b** Biased classification results of ResNet-10 on ImageNet-LT

However, the above assumption might not always hold true as certain classes might be well-represented even by a small number of data. To support our claim, we conduct a simple experiment, where we train a ResNet-10 model on a long-tailed dataset (ImageNet-LT (Deng et al., 2009)) in the conventional manner (cross-entropy loss) and then test the model's class-wise performance on the test set. The result is shown in Fig 1. Although the accuracy of different classes tends to decrease as the number of training samples decreases, it does not monotonically decrease at all and there is significant fluctuation: there are many classes with less training samples that have higher accuracy than some classes with much more samples. For example, the accuracy of class#2 (1246 training samples) is 0.45 while that of class#239 (150 training samples) is 0.95. This implies that not all tail classes are necessarily difficult and therefore do not obviously require high focus. This makes us rethink if the class-wise frequency is actually the best clue to tackle the long-tail problem.

We believe it is more reasonable to use difficulty instead of the frequency as a clue to determine the weights. There are some previous researches that have proposed to use difficulty such as Focal loss (Lin et al., 2017). Focal loss is a cost-sensitive learning strategy that measures instance-level difficulty (also known as hard-mining) and accordingly assigns higher weights to harder samples. Researchers have widely used focal loss for long-tailed recognition problems as tail classes have higher proportion of hard instances, and therefore giving more weight to such instances should presumably give more weight to the tail classes. But, unfortunately, that does not hold true because even though the *proportion* of hard instances in tail classes is higher than that in head classes, the *absolute number* of hard instances in the tail classes is still lower. We provide more evidence and in-depth analysis in Sect. 4.1.6.

Therefore, we think it is more sound to use class-level difficulty instead of instance-level. Recently, we (Sinha et al., 2020) proposed a simple yet effective method for measuring the class-wise difficulty. We used the difficulty measure for dynamically assigning weights to the loss of training samples and showed the effectiveness of the weighted loss in mitigating the class-imbalance problem in classification tasks.

This paper extends the previous paper and provides more through analysis, extensive experimental results, and deeper insights on the class-wise difficulty based method. The key differences of this work with the previous work (Sinha et al., 2020) and the contributions of this work is four-fold.

First, we extended the experiments to include long-tailed object detection and instance segmentation on LVIS. This makes the evaluation more complete, showing that the proposed difficulty measure thoroughly applies to a variety of computer vision tasks including image classification, object detection, instance segmentation, and video action recognition.

Second, we provide detailed analysis on the behavior of the class-wise method in comparison to focal loss (a typical example of instance-level method) to better understand the characteristics of these methods (in Sect. 4.1.6). We argued why the proposed class-level difficulty empirically performs better in the experiments. To the best of our knowledge, this is the first work to deeply analyze the different behavior of class-difficulty-based methods and instance-difficulty-based methods.

Third, we show that the class-wise difficulty measure can be also applicable to sampling technique, which is another most popular strategy in mitigating the class-imbalance problem along with weighted loss. This is important because weighting and sampling are the two most popular technique for mitigating class imbalance, and they are sometimes used in different contexts. For example, different sampling techniques are employed in the decoupled learning (Kang et al., 2020). We show the effectiveness of CDB-S in this framework as well as in the end-to-end training framework in Sects. 4.1.3 and 4.1.5.

Finally, we provide more justification and reasoning on the way of dynamical update of the focusing parameter $\tau$ presented in (Sinha et al., 2020). We pay attention to the growth speed of $\tau$ in relation to the imbalance in performance which we call *bias*. We designed four different variants to represent different growth speed and empirically tested their performance. As a consequence of experiments, we found the preferable growth rate of $\tau$ has some relationship with the imbalance injected in training data.

## 2 Related Works

Long-tail in datasets is a very fundamental problem that is encountered quite frequently, especially when the final goal is real-world deployment. Most prior researches in this field can be fairly categorized into three domains: (1) Data re-sampling techniques (Chawla et al., 2002; Barua et al., 2014; Liu et al., 2009), (2) Cost-sensitive learning methods (Lin et al., 2017; Cui et al., 2019; Tan et al., 2020; Cao et al., 2019; Sinha et al., 2020) and (3) Metric learning and knowledge transfer (Liu et al., 2019; Song et al., 2016; Huang et al., 2020).

Data re-sampling techniques try to create balance among the classes during the data pre-processing step. Such balance is created by oversampling from the tail classes or under-sampling from the head classes. Oversampling methods generally generate new data samples for the classes either by replication or synthetic approaches (Chawla et al., 2002; Han et al., 2005). Naive oversampling by data replication tends to cause overfitting (He & Garcia, 2009; Fernández Hilario et al., 2018), while synthetic generation must ensure that distinct features of the tail classes reflect in the generated data. Under-sampling methods (Liu et al., 2009), on the other hand, reduce data from the head classes by various approaches. However, random undersampling might cause the model to miss significant concepts related to the head classes (He & Garcia, 2009; Fernández Hilario et al., 2018). Therefore, both oversampling and undersampling need to be employed carefully to be effective.

Cost-sensitive learning penalizes the DNN model differently for different training samples. This is achieved by assigning weight values to each sample, which is then used to scale the penalty incurred for prediction mistakes on the samples. The main motivation of this research field is to search for an efficient and effective weight assignment system. To solve class-imbalance, the typical approach is to assign higher weights to the samples of the tail classes while lowering weights for the head class samples. The most common and simple weighting strategies involve weight assignment by inverse of class frequencies or a smoothed version of the function. Recently, Yin et al. (Cui et al., 2019) proposed a weight assignment strategy that uses the effective number of samples for each class. Jamal et al. (Jamal et al., 2020) claimed that class-imbalanced recognition problem can be considered a special case of domain adaptation and therefore, proposed a weighting strategy motivated from domain adaptation. Many other works such as balanced group softmax (Li et al., 2020) and equalization loss (Tan et al., 2020) try to provide better weight assignment systems, and the primary attribute used for deciding the weights is the class-wise frequency. For many real datasets, it has been seen that the state-of-the-art (SOTA) cost-sensitive learning techniques outperform the SOTA sampling techniques. However, Kang et al. (Kang et al., 2020) shows that by decoupling DNN learning into representation learning and classification, it is possible to improve the performance of the sampling techniques but such a procedure becomes time-consuming due to use of multiple training stages. Most of the above-mentioned sampling and weight-assignment techniques assume that the tail classes are always under-represented and therefore, the most difficult classes to be learned by the DNN. However, such an assumption might not hold true in cases, where a class is sufficiently well-represented even with a small number of samples.

An alternate way of weighing samples is to assign weights to each sample based on it's difficulty. Samples with higher difficulty are weighed higher compared to the easier ones. Such an approach was initially proposed for the task of dense object detection in form of focal loss (Lin et al., 2017), however recent works have extended it to long-tailed recognition problems as well. As the tail class samples are expected to be the most difficult samples, it is assumed that such a difficulty-based weighting would eventually put more weight on the tail classes. But due to the relatively large population of the head classes, they tend to have higher absolute number of hard (high difficulty) samples than the tail classes. As a result, the above approach ends up putting more weight on the head classes, resulting in biased results, evidence of which will be shown in Sect. 4.1.6. However, such an approach interests us due to the novel concept of using difficulty over frequency for weight assignment. Therefore, in this work, we propose to use the class-wise difficulty for balancing the data distribution, while addressing the shortcoming of focal loss.

Recently metric learning and knowledge transfers approaches have become popular in the field of long-tail recognition. Metric learning (Huang et al., 2020; Song et al., 2016) approaches try to find an appropriate embedding function that embeds data into the feature space, while preserving certain inter-data relationships. Contrastive embedding uses pairs of data samples to minimize the intra-class feature distance, while maximizing the inter-class distance. Triplet loss (Huang et al., 2020) uses triplets of data samples for similar purpose, where one of the sample is considered an anchor. However, for large datasets, it is computationally inefficient to select all possible pairs or triplets for training. In such cases, the performance of these metric learning approaches depends largely on the choice of pair or triplet selection procedure, which is a different research domain. Some works (Liu et al., 2019; Wang et al., 2020, 2017) have proposed the knowledge transfer from head classes to tail classes either using meta learning or an external memory module. Though an effective approach, such modules are generally not end-to-end trainable and designing them might be a time-consuming, expensive and non-trivial task. That is why, this work focuses mainly on data re-sampling and cost-sensitive learning approaches.

## 3 Proposed Method

### 3.1 Class-wise Difficulty

Human beings use the metric 'difficulty' to provide qualitative descriptions of tasks e.g., "this home-work is difficult" or "playing soccer is easy". DNN models behave quite similarly, where they find some tasks harder to perform compared to others. Likewise, in long-tailed recognition problems, the models find some classes relatively easier to learn. As stated in Sect. 2, most prior works assume that the head classes are always easier to learn for the model than the tail classes, which might not always hold true. Such assumptions can be avoided, if we can find a way to quantitatively determine the metric difficulty. However, adding a quantitative value to a qualitative metric is one issue. Lin et al. (Lin et al., 2017) tried to solve that by proposing a cost-sensitive learning using 'focal loss', where each training instance is assigned a weight based on its difficulty. Focal loss calculates the difficulty of an instance using the DNN model's prediction output for the instance. If an instance $s$ belongs to class $y_c$ among a given set of class labels $Y = \{y_1, y_2, \ldots, y_N\}$, then the DNN model's output for $s$ can be represented as $\{p_{y_1,s}, p_{y_2,s}, \ldots, p_{y_N,s}\}$ where $p_{y_n,s}$ is the model's prediction confidence corresponding to the class $y_n$ for sample $s$. Focal loss calculates the difficulty of instance $s$ as $d_s = (1 - p_{y_c,s})$ and computes the weight for sample $s$ as $d_s^\gamma$, where $\gamma$ is a hyper-parameter called 'focusing parameter'. Higher value of $d_s$ signifies higher sample difficulty and leads to higher sample weight.

Even though focal loss successfully added quantitative value to the metric 'difficulty', it fails to solve long-tailed recognition problems due to reasons stated in Sect. 2. Evidence of this will be presented in Sect. 4.1.6. We hypothesize that the drawback of focal loss appears because it uses the difficulty of each instance. However, for long-tailed problems, it is more important to determine the difficulty of the classes than that of the instances. Therefore, we propose to use the class-wise difficulty to solve the imbalance. But as stated before, there is no direct way to add a quantitative value to such qualitative metric. Hence, we find a novel way to do so.

Humans tend to call a task difficult if they perform poorly in it. For example, Mom : "Son, why did you get only 30 in Mathematics?"; Son : "Because the paper was very difficult". We use a similar concept to quantify the class-wise difficulty as perceived by the DNN model. We measure the performance of the model for each class and use it to determine the perceived difficulty of each class. The performance metric used can vary based on the tasks. For classification tasks, the performance metric is the class-wise classification accuracy while for object detection tasks, it can be average precision. Our approach is proposed using performance metrics where higher means better, but it can be easily adapted to other metrics as well. For simplicity, we limit our explanation here to classification tasks. For multi-class classification task with $N$ classes and $M$ data samples, the classification accuracy of class $c$ can be calculated as

$$A_c = \frac{m_c}{M_c}, \qquad (1)$$

where $M_c$ is the total number of samples belonging to class $c$ and $m_c$ is the number of samples belonging to class $c$ that have been classified correctly. Therefore, $\sum_{c'=1}^{N} M_{c'} = M$ and $m_{c'} \leq M_{c'}, \quad \forall c' \in 1, 2, \ldots, N$. We use the classification accuracy of class $c$, $A_c$, for computing the difficulty of class $c$ using

$$d_c = 1 - A_c. \qquad (2)$$

For computing the class-wise difficulty, we use a class-balanced subset of the dataset, which is different from both the train and test set. This is done to prevent over-fitting to either train or test set. The separate subset used to measure difficulty can also be used as validation or development set. Another interesting observation about human learning and DNN learning is that as the learning progresses, the difficulty of a task or a class reduces gradually. That is why, we propose to measure the class-wise difficulty dynamically during the training period. For dynamic calculation, we compute the difficulty scores after every $t$ epochs. In our experiments, we use $t = 1$. Therefore, the dynamic difficulty of a class $c$ can

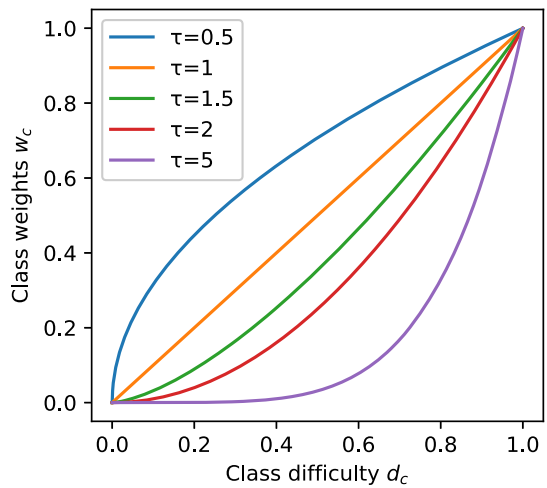

**Fig. 2** Effect of varying the value of $\tau$ on the class-difficulty based weight assignment. Higher value of $\tau$ puts heavier weights on the samples of the harder classes, while relatively reducing weights on the easier class samples

be computed using

$$d_{c,it} = 1 - A_{c,it}, \quad (3)$$

where $A_{c,it}$ is the model's classification accuracy on class $c$ after $it$ epochs and $i \in \{1, 2, ..., \frac{T_{max}}{t}\}$. $T_{max}$ is determined by the total number of epochs used for training.

### 3.2 Difficulty-Based Weight Assignment

We propose to use the computed class-wise dynamic difficulty to dynamically assign weights to the different training samples. The obvious way to do that is by assigning higher weights to the samples of the class that have high difficulty scores. Therefore, we propose a dynamic weight assignment using

$$w_{c,it} = (d_{c,it})^\tau, \quad (4)$$

where $d_{c,it}$ is computed from Eq. 3 and $\tau$ is a hyper-parameter that controls how much we emphasize more difficult classes. Using Eq. 4, if the minority class is actually a very difficult class, then all the samples of that class will get high weights, resulting in overall high weight for the minority class. Therefore, this approach overcomes the drawback of focal loss.

The effect of changing the value of $\tau$ on the weight distribution is visualized in Fig. 2.

It is important to find out a reasonable value of $\tau$ for the fast training convergence and balanced performance. As shown in the figure, selecting too high value of $\tau$ (e.g., $\tau = 5$) excessively down-weights easier classes, which results in too slow training progress or too skewed performance, while selecting too low value for $\tau$ results in almost similar biased performance as the unweighted case ($\tau=0$ reduces to equal weights for all samples).

It is difficult, however, to find out a perfect value for $\tau$ because it may change according to the characteristics of datasets such as the level of class-imbalance and the difference of each class's 'difficulty'. Moreover, the 'difficulty' of each class usually changes dynamically along with the training progress as explained above.

That is why we propose an approach to automatically compute the reasonable value of $\tau$, taking the dynamic difficulties in the training process into account. Based on the characteristics of $\tau$ observed in Fig. 2, we think it is better to use bigger $\tau$ if we need to emphasize the difficult classes more compared to easy classes, and use smaller $\tau$ if we do not have to emphasize the difficult classes very much. It is reasonable to assume that we need to emphasize the difficult classes more when there is severe imbalance in performance on different classes. Therefore, it is necessary to quantify this imbalance in performance. We call this imbalance in performance "*bias*", and design it such that it satisfies the following properties. (1) It becomes bigger if certain class(es) performs significantly better (or worse) than other classes. (2) It becomes 0 if the performance is perfectly balanced, namely the accuracy of all the classes are the same. (3) It does not become a negative value. Mathematically, we define it as follows.

$$bias_{it} = \max\left(\frac{\max_{c=1}^{N} A_{c,it}}{\min_{c'=1}^{N} A_{c',it} + \epsilon} - 1, 0\right). \quad (5)$$

$\epsilon$ is a small positive value used to avoid errors in cases where $\min_{c'=1,2,...,N} A_{c',it} = 0$. $\epsilon$ also helps to keep $bias_{it}$ upper bounded by $\epsilon^{-1} - 1$, which might otherwise explode to $\infty$. For our experiments, we use $\epsilon = 0.01$.

Next, we define $\tau$ as a function of the *bias* as

$$\tau_{it} = f(bias_{it}). \quad (6)$$

Note that the value of $\tau$ is computed and updated after every $t$ epochs using the model's performance bias.

Now, we want to design $f$ such that the value of $\tau$ increases when the *bias* in performance increases (imbalanced performance), resulting in much heavier weights for the difficult classes. To satisfy this, the function $f$ should monotonically increase with $bias_{it}$. The question is how much the increase speed should be. Here we design four different variants of $f$ that represent different growth speed of $\tau$ with respect to *bias* as shown in Fig. 3, and give empirical analysis of these variants in Sect. 4.1.4.

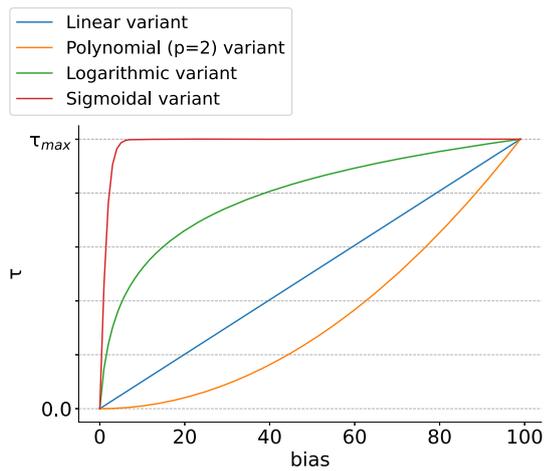

**Fig. 3** Four different variants for dynamic $\tau$ calculation

- Linear $f$ : The linear variant of $f$ computes the value of $\tau$ as

$$\tau_{it} = \frac{bias_{it}}{B_M} T_M, \tag{7}$$

where $B_M$ is the upper bound of bias computed by Eq. 5, which is $\epsilon^{-1} - 1$ and $T_M$ is the predefined upper bound of the value of $\tau$. For our experiments, we use $T_M = 5$. The linear variant maintains a uniform growth speed of $\tau$ with respect to the bias.

- Polynomial $f$ : The polynomial variant of $f$ computes $\tau$ using

$$\tau_{it} = \left(\frac{bias_{it}}{B_M}\right)^p T_M, \tag{8}$$

where $p$ can assume positive integers greater than 1. $p = 1$ gives the linear variant of $f$. The polynomial variant has very slow increase of $\tau$ when the $bias$ is low. Generally in long-tailed learning, the $bias$ is relatively low in the early stages in the training, and it gradually increases as the training progresses. Therefore, this variant tends to emphasize all the classes almost uniformly in the initial stages of the training, in which the $bias$ tends to be small. However, when the $bias$ gets higher and higher in later stage of training, the polynomial variant quickens the pace of $\tau$ growth, resulting in faster increase of emphasis on difficult classes.

- Logarithmic $f$ : The logarithmic variant of $f$ computes $\tau$ using

$$\tau_{it} = -\frac{T_M}{\log(B_M + 1)} \log(bias_{it} + 1). \tag{9}$$

The logarithmic variant set $\tau$ higher even when the $bias$ is lower, resulting in fast increase of emphasis on the hard classes at the early stages of training.

- Sigmoidal $f$ : This variant of $f$ computes $\tau$ as

$$\tau_{it} = 2 \cdot T_M \cdot \text{sigmoid}(bias_{it}) - T_M$$
$$= T_M\left(\frac{2}{1 + e^{-(bias_{it})}} - 1\right). \tag{10}$$

The sigmoidal variant follows a similar pattern as the logarithmic one but increases $\tau$ much faster, resulting in very early increase in emphasis on the difficult classes and then maintaining the high emphasis unless the performances on different classes are very-well balanced.

### 3.2.1 Class-wise Difficulty-Based Weighted Loss

For explanation of class-wise difficulty-based weighted (CDB-W) loss, we use the most conventional cross-entropy loss function and focal loss. However, the application of CDB weighting is not limited to only these losses but can be easily applied to other loss functions such as mean-squared error loss, hinge loss etc.

Cross-entropy loss is used in most classification tasks. For a $N$-class classification task, if a data sample $s$ of a class $c$ is forward propagated through the classifier in $k^{th}$ epoch, then the model outputs a probability distribution over the $N$ classes given as $p_{s,k} = \{p_{s,1,k}, p_{s,2,k}, \ldots, p_{s,N,k}\}$. $p_{s,c',k}$ is the prediction probability of sample $s$ belonging to class $c'$ during $k^{th}$ epoch. The cross-entropy (CE) loss for sample $s$ during $k^{th}$ epoch is computed as

$$CE_{s,k} = -\log(p_{s,c,k}), \tag{11}$$

where sample $s$ belongs to class $c$. In similar situation, class-wise difficulty-based weighted cross-entropy (CDB-W-CE) loss calculates the loss as

$$CDB\text{-}W\text{-}CE_{s,k} = -w_{c,k-1} \log(p_{s,c,k})$$
$$= -(d_{c,k-1})^{\tau_{k-1}} \log(p_{s,c,k}), \tag{12}$$

where $d_{c,k-1}$ is computed using Eq.3 before $k^{th}$ epoch and $\tau_{k-1}$ is the value of $\tau$ updated before $k^{th}$ epoch. Our CDB weighting can be applied to focal loss as well. For the above sample $s$, focal loss (FL) is computed as

$$FL_{s,k} = -(1 - p_{s,c,k})^\gamma \log(p_{s,c,k}), \tag{13}$$

and class-wise difficulty-balanced weighted focal loss (CDB-W-FL) is computed as

$$CDB\text{-}W\text{-}FL_{s,k} = -(d_{c,k-1})^{\tau_{k-1}}(1 - p_{s,c,k})^\gamma \log(p_{s,c,k}). \tag{14}$$

### 3.2.2 Class-wise Difficulty-Based Sampling Method

For long-tailed recognition, sampling methods aim to create a balance in the data distribution during the pre-processing step. Generally, they achieve this by sampling training data from different classes using unequal probabilities. However, all prior sampling methods assume that the tail classes are always under-represented and difficult to be learned. As argued before, that might not always hold true. That is why, we propose a sampling method using our class-wise difficulty scores called class-wise difficulty-based sampling method (CDB-S). CDB-S samples a training data from class $c$ in the $k^{th}$ epoch using probability

$$p_{c,k} = \frac{w_{c,k-1}}{\sum_{c'=1}^{N} w_{c',k-1}} = \frac{(d_{c,k-1})^{\tau_{k-1}}}{\sum_{c'=1}^{N} (d_{c',k-1})^{\tau_{k-1}}}, \quad (15)$$

where $d_{c',k-1}$ is the difficulty of class $c'$ computed using Eq.3 after $k-1$ epochs. By this method, the sampler samples a training data from a difficult class more frequently compared to an easy class, helping the model to learn from both easy and difficult classes in a balanced way.

## 4 Evaluation

Long-tail is a fundamental issue encountered in multiple recognition tasks such as image classification, object detection, instance segmentation, video action classification etc. We believe that our proposed CDB weighted loss and sampling method will be effective for any such tasks. Therefore, to verify the task-agnostic performance of the proposed method, we conduct experiments on 4 different tasks namely image classification, object detection, instance segmentation and video-action classification. For each of these tasks, we compare our methods with various SOTA methods.

### 4.1 Image Classification

#### 4.1.1 Datasets

**MNIST** MNIST (LeCun, 1998) is a popular image classification dataset, where the task is to classify gray-scale images of hand-written digits into one of the ten classes of digits (0-9). There is no inherent class-imbalance in this dataset. However, we generate a class-imbalanced subset of the dataset following same setup as (Ren et al., 2018). We build a training subset by using only 5000 samples from 2 classes (4 and 9). The imbalance was injected into the subset by selecting unequal number of training samples for the classes. The number of samples for each class was determined by the value of 'head class ratio'. If 'head class ratio' is $x$, then number of training samples in the head class is $x * 5000$ and that in tail class is $(1 - x) * 5000$. This procedure is same as Ren et al. (2018). For our experiments, we choose the class '4' as our head class and the 'head class ratio' is selected as 0.99 or 0.995. To facilitate class-wise difficulty calculation, we build a balanced subset from the data using 500 samples for both class '4' and '9'. This subset is different from the training subset and can also be used for validation purposes. For testing, we also create a balanced test subset using 800 samples for both the classes, which is different from the 2 above-mentioned subsets. For various values of 'head class ratio', we compare the performance of our proposed methods with other SOTA methods.

**CIFAR100-LT** CIFAR100 (Krizhevsky, 2009) is another popular image classification dataset, where there are 100 different classes such as beaver, shark, roses etc. and the task is to classify tiny RGB images into the corresponding correct classes. Similar to MNIST, this is also an inherently class-balanced dataset. However, long-tail can be injected into the dataset using techniques as proposed in (Cui et al., 2019), to create a long-tailed version popularly called CIFAR100-LT. The balanced dataset has 500 training images and 100 testing images for each of the 100 classes. We create a small validation subset using only 50 samples per class from the training set for difficulty calculation. The remaining 450 images per class are used to build an imbalanced training set. In the imbalanced training set, the number of selected training samples for a class $c$ is determined as $n_c = n_{\max} \mu^{\frac{c-1}{N-1}}$, where $n_{\max} = 450$ and $N$ is total number of classes, which is equal to 100. For testing, we use the same balanced test set as provided. We vary the value of $\mu$ to increase or decrease class-imbalance in training set and evaluate the performance of our proposed methods for various $\mu$.

**ImageNet-LT** ImageNet is a large-scale image classification dataset, which comprises of 1000 image categories or classes. This dataset is not inherently class-imbalanced. However, a long-tailed version of the dataset has been constructed in (Liu et al., 2019). It comprises a long-tailed training set of 115,800 images, where the number of samples per class varies from 5 to 1280. Separate balanced validation and test sets are also provided. We use the validation set for difficulty calculation and the test set for evaluation.

#### 4.1.2 Implementation Details

For MNIST (LeCun, 1998), we train a LeNet-5 (LeCun et al., 1998) on imbalanced training set for 100 epochs using a batch size of 100. LeNet-5 is used to keep the experimental setup exactly same as (Ren et al., 2018). SGD optimizer is used with a weight decay of 0.0005, momentum of 0.9 and an initial learning rate of 0.001. The training is done on a single NVIDIA GeForce GTX 1080 GPU.

For CIFAR100-LT, we follow similar training strategy as (Cui et al., 2019). We train a ResNet-32 (He et al., 2015) for 200 epochs using a batch size of 128 on 4 NVIDIA Titan X GPUs. We use SGD optimizer with momentum 0.9, weight decay 0.0001 and an initial learning rate of 0.1, which is scaled down after 160 and 180 epochs.

For ImageNet-LT, we use a ResNet-10 and train it for 100 epochs using batch size 512. As before, SGD optimizer is used with momentum 0.9, weight decay 0.0005 and an initial learning rate of 0.2, which is subjected to cosine decay. We used two NVIDIA GeForce GTX 1080 Ti GPUs for ImageNet-LT experiments.

For each of the datasets, we use a balanced validation dataset for calculating the class-wise difficulty scores, which is important for our proposed CDB methods. The trained models are evaluated on the balanced test sets. All implementations are done in PyTorch.

### 4.1.3 Comparison with SOTA

To evaluate the effectiveness of our proposed methods, we compare its performance with various methods i.e., (1) Cross-entropy (CE) loss, (2) Class-frequency based sampling (CFS) (Mikolov et al., 2013), (3) Focal loss (Lin et al., 2017), (4) Class-balanced (CB) loss (Cui et al., 2019), (5) Equalization loss (EQL) (Tan et al., 2020), (6) Learning to reweighting (L2RW) (Ren et al., 2018), (7) Meta-weight Net (Shu et al., 2019), (8) LDAM-DRW (Cao et al., 2019), (9) OLTR (Liu et al., 2019), and (10) Class-aware sampling (Kang et al., 2020). Recently (Kang et al., 2020) showed that it is effective to decouple representation learning and classifier learning. However, we do not employ this 2-stage training strategy in this section in order to assess the pure effectiveness of the class-wise difficulty idea in end-to-end training. As such, we compare the proposed method with only end-to-end single-stage training methods. The effectiveness of CDB methods in decoupled learning is assessed later in Sect. 4.1.5. The comparison is done over 3 image datasets and in each case the results were calculated on a balanced test set. For each of the datasets, we report the results of CDB methods using sigmoidal variant of dynamic $\tau$ calculation because we found it performs reasonably well in most cases as we will show in Sect. 4.1.4.

As can be seen from Table 1, our CDB weighted loss and sampling methods perform better than most other SOTA methods for all the datasets. The results show that the CDB weight assignment is an effective way to design a weighted loss or sampling method for solving class-imbalance. In most cases, sampling techniques give lower performance than most weighted loss approaches, which is evident from the results of class-frequency based sampling (CFS) in Table 1. But, CDB-S sampling not only outperforms class-frequency based sampling (e.g., by 3.04% for CIFAR100-LT imbalance 100) but also consistently provides better results than many SOTA weighted loss techniques such as CB loss, L2RW, EQL (e.g., **1.43**% better than EQL (Tan et al., 2020) for CIFAR100-LT imbalance 100). Another interesting thing is that using our CDB weight assignment with focal loss helps to improve the performance of focal loss significantly (e.g., **3.20**% for ImageNet-LT).

### 4.1.4 Effect of $\tau$

We conduct an experiment to study the effect of $\tau$ on the performance of our proposed method. For the experiment, we train a ResNet-32 on CIFAR100-LT using both CDB-W-CE loss and CDB-S sampling with different formulations for $\tau$. We use both fixed value as well as dynamic calculation for $\tau$. For fixed value $\tau$, we increase $\tau$ in the range of [0.5, 5] and study the corresponding changes in the performance. For dynamic calculation, we experiment with all the 4 proposed variants. The results are listed in Tables 2 and 3.

Tables 2 and 3 show that, given a fixed imbalance value, as we increase the value of $\tau$, the performance of both CDB-W-CE and CDB-S initially improves, however after a certain point it starts to drop. For higher imbalance, the best performing $\tau$ for both CDB-W-CE and CDB-S is generally higher than that for low imbalance, which is reasonable consequence. The important observation is that the optimal value for fixed $\tau$ varies with both the amount of imbalance and the dataset. For example, CDB-W-CE with $\tau = 1.5$ performs best for CIFAR-LT with imbalance 50, while $\tau = 1.0$ performs best for imbalance 200. Also, for imbalance 99 in MNIST, CDB-W-CE with $\tau = 2.0$ works best while for CIFAR-LT with imbalance 100, $\tau = 1.5$ works best. As easily imagined, it is time-consuming to manually find out the reasonable value of $\tau$.

Due to the difficulty in searching for a single optimum $\tau$, we proposed 4 different variants to dynamically calculate $\tau$. Evidently for both CDB-W-CE and CDB-S, the sigmoidal variant performs better than the others in most cases and its performance is almost always close to the best of fixed $\tau$ or sometimes even better than the best fixed $\tau$ because these variants can dynamically change the value of $\tau$ in accordance with the training progress. This signifies that giving high emphasis on difficult classes from as early stage of training as possible benefits the model especially when training data is highly imbalanced. It is also seen that slower increase in emphasis can sometimes give comparable performance in low-imbalance situations (CDB-S with linear $\tau$ for CIFAR100-LT imbalance=10), which supports that having moderate focus on the hard classes can be enough when the imbalance in data is not high. Considering the consistently good performance of the sigmoidal variant in various settings, we use it in the remainder of the experiments unless otherwise stated.

**Table 1** Classification errors (%) of different methods for solving class-imbalance in 3 different image datasets

| Dataset | MNIST | | CIFAR100-LT | | ImageNet-LT |
| --- | --- | --- | --- | --- | --- |
| Imbalance | 99 | 199 | 10 | 100 | 256 |
| CE loss | 8.59 ± 0.41 | 14.35 ± 1.10 | 44.35 | 61.79 | 65.20 |
| Focal loss (Lin et al., 2017) | 6.57 ± 0.61 | 11.45 ± 0.74 | 48.05 | 61.59 | 69.50 |
| CFS (Mikolov et al., 2013) | 6.10 ± 0.46 | 10.59 ± 0.65 | 42.66 | 61.07 | – |
| CB loss (Cui et al., 2019) | 5.88 ± 1.20 | 8.61 ± 1.11 | 42.11 | 60.40 | – |
| EQL (Tan et al., 2020) | 2.60 ± 0.33 | **3.71 ± 0.41** | 41.68 | 59.46 | 63.56 |
| L2RW (Ren et al., 2018) | 2.63 ± 0.65 | 3.94 ± 1.23 | 46.27 | 59.77 | - |
| LDAM-DRW (Cao et al., 2019) | – | – | 41.29 | 57.96 | – |
| Meta-weight Net (Shu et al., 2019) | – | – | 41.54 | 57.91 | – |
| OLTR (Liu et al., 2019) | – | – | – | – | 64.40 |
| Class-aware sampling (Kang et al., 2020) | – | – | – | – | 63.00 |
| CDB-S (Ours) | 2.80 ± 0.35 | 3.93 ± 0.65 | 41.39 | 58.03 | 62.30 |
| CDB-W-FL (Ours) | 4.12 ± 0.32 | 5.64 ± 0.51 | 42.58 | 59.69 | 66.30 |
| CDB-W-CE (Ours) | **2.39 ± 0.41** | **3.71 ± 0.27** | **41.26** | **57.41** | **61.50** |

The best results for each dataset and each imbalance values are highlighted in bold
Note that the imbalance values are computed as the ratio of the number of training samples in the most frequent class to that of the least frequent class

**Table 2** Classification error (%) on 3 image datasets using CDB-W-CE loss with different values of $\tau$

| | | MNIST | | CIFAR100-LT | | | | | ImageNet-LT |
| --- | --- | --- | --- | --- | --- | --- | --- | --- | --- |
| Imbalance | | 99 | 199 | 10 | 20 | 50 | 100 | 200 | 256 |
| Fixed $\tau$ | 0.5 | 2.59 ± 0.44 | 4.03 ± 0.41 | 41.71 | **45.40** | 53.87 | 58.74 | 62.79 | 63.00 |
| | 1.0 | 2.49 ± 0.57 | 3.94 ± 0.36 | **40.53** | 46.52 | 53.55 | 58.33 | **62.01** | 62.90 |
| | 1.5 | 2.31 ± 0.38 | **3.54 ± 0.25** | 41.32 | 47.26 | **52.91** | **57.30** | 62.37 | 62.70 |
| | 2.0 | **2.23 ± 0.34** | 3.65 ± 0.41 | 41.35 | 47.00 | 53.06 | 57.56 | 62.97 | **62.30** |
| | 5.0 | 2.51 ± 0.41 | 3.78 ± 0.43 | 45.52 | 48.33 | 54.55 | 59.38 | 63.19 | 63.50 |
| Dynamic $\tau$ | Linear | 3.06 ± 0.32 | 3.56 ± 0.45 | 42.16 | 46.68 | 54.32 | 58.40 | **62.24** | 62.80 |
| | Poly | 3.25 ± 0.43 | 3.51 ± 0.46 | 42.15 | 47.64 | 54.55 | 58.59 | 62.66 | 62.60 |
| | Log | 2.75 ± 0.37 | **3.37 ± 0.36** | 42.32 | 47.73 | 54.70 | 58.77 | 63.39 | 62.10 |
| | Sigmoid | **2.39 ± 0.41** | 3.71 ± 0.27 | **41.26** | **44.94** | 53.18 | **57.41** | 62.28 | **61.50** |

For this experiment, we used both fixed $\tau$ and dynamic $\tau$. For each imbalance values, we have emboldened the best results for both fixed $\tau$ and dynamic $\tau$

### 4.1.5 Results in Decoupled-Learning

Recently Kang et. al. (Kang et al., 2020) proposed that decoupling the representation and classifier learning can be very effective for long-tailed recognition. For that they introduced 2-stage training procedure, where the first stage learns a powerful feature extractor and the latter stage balances the classifier. For the second stage, class-aware sampling is generally used. In class-aware sampling, each class is assigned a fixed and equal sampling probability.

We investigated the effect of our proposed class-difficulty based methods in both stages of decoupled learning. Table 4 shows the results for 2 classifier balancing methods in the 2nd stage of the decoupled learning - classifier retraining (cRT) and learnable weight scaling (LWS).

As seen in the table, for the 1st stage training, CDB-based methods (both CDB-W-CE and CDB-S) are clearly better than vanilla cross entropy loss. This indicates that the proposed CDB-based methods are effective in learning powerful representation.

For the 2nd stage, the sampling-based methods (CDB-S and Class-aware sampling) are better than weighting-based method (CDB-W-CE). Among the sampling-based methods, CDB-S is better than Class-aware sampling because it reflects the performance of the classifier on-the-fly, which can dynamically change as the training progresses.

Based on these observations, we conclude that the best choice in the decoupled learning is to use CDB-based methods in the 1st stage to learn better representations, and to use CDB-S in the 2nd stage to better re-balance the classifier.

**Table 3** Classification error (%) on 3 image datasets using CDB-S loss with different values of $\tau$

| Imbalance | | MNIST 99 | 199 | CIFAR100-LT 10 | 20 | 50 | 100 | 200 | ImageNet-LT 256 |
|---|---|---|---|---|---|---|---|---|---|
| Fixed $\tau$ | 0.5 | 3.24 ± 0.37 | 5.11 ± 0.56 | 41.96 | 47.59 | **53.95** | 59.26 | 63.17 | 63.30 |
| | 1.0 | **2.71 ± 0.34** | 4.56 ± 0.48 | **41.63** | 47.37 | 54.81 | 59.54 | 63.20 | 63.10 |
| | 1.5 | 2.88 ± 0.56 | 4.17 ± 0.46 | 42.32 | 47.41 | 54.41 | 59.12 | 63.87 | **62.70** |
| | 2.0 | 3.03 ± 0.44 | **3.86 ± 0.43** | 42.09 | **47.30** | 54.83 | 59.31 | 63.52 | **62.70** |
| | 5.0 | 3.43 ± 0.43 | 4.24 ± 0.58 | 46.29 | 49.33 | 54.37 | **58.70** | **62.55** | 65.50 |
| Dynamic $\tau$ | Linear | 2.97 ± 0.24 | 4.19 ± 0.51 | 41.51 | 47.05 | 53.58 | 58.94 | 63.19 | 62.80 |
| | Poly | 3.12 ± 0.26 | 4.11 ± 0.34 | 42.40 | 48.53 | 54.13 | 59.31 | 62.71 | **62.30** |
| | Log | 2.91 ± 0.34 | 4.00 ± 0.36 | 42.65 | 47.10 | 53.74 | **57.79** | 63.00 | 62.70 |
| | Sigmoid | **2.80 ± 0.35** | **3.93 ± 0.65** | **41.39** | **46.08** | **53.37** | 58.03 | **62.26** | **62.30** |

For each imbalance values, we have emboldened the best results for both fixed $\tau$ and dynamic $\tau$

**Table 4** Classification error (%) on CIFAR100-LT for decoupled learning

| 2nd stage training methods | | Imbalance: 10 1st stage training methods CE loss | CDB-W-CE | CDB-S | Imbalance: 100 1st stage training methods CE loss | CDB-W-CE | CDB-S |
|---|---|---|---|---|---|---|---|
| cRT | Class-aware sampling | 41.28 | 40.99 | <u>40.97</u> | 57.06 | <u>56.74</u> | 56.87 |
| | CDB-W-CE | 41.47 | <u>41.33</u> | 41.35 | 57.44 | 57.38 | <u>57.36</u> |
| | CDB-S | **41.05** | **40.89** | 40.92 | 56.49 | **56.24** | 56.29 |
| LWS | Class-aware sampling | 40.64 | <u>40.42</u> | 40.48 | 56.68 | <u>56.41</u> | 56.45 |
| | CDB-W-CE | 41.11 | <u>41.02</u> | 41.06 | 57.20 | <u>56.88</u> | 56.99 |
| | CDB-S | ***40.44*** | ***40.16*** | 40.21 | 56.26 | <u>**56.11**</u> | 56.12 |

The second row corresponds to the stage-1 learning methods while the second column lists the second stage learning methods. Within each block, the results of best 1st stage methods (the best in each row) are highlighted with <u>underline</u>, and the results of best 2nd stage methods (the best in each column) are highlighted with **bold**. The best in each block is shown in *italic*

Overall, the result supports the effectiveness of the newly proposed CDB-S especially in the decoupled learning scenario.

### 4.1.6 Comparison to Focal Loss

Here we make a detailed analysis of how CDB-W-CE helps to solve the drawbacks of the focal loss.

First, we provide evidence for the drawbacks of focal loss as stated in Sect. 2 by conducting a simple experiment. Based on the number of training instances of each class ($N_c$), we divide the classes of CIFAR100-LT (imbalance=200) into many-shot ($N_c > 100$), medium-shot ($100 \geq N_c > 20$) and few-shot ($20 \geq N_c$) classes. While training a ResNet-32 on CIFAR100-LT using focal loss, we track the average number of hard instances in each of the three subsets of classes as shown in Fig. 4. We classify an instance as hard if it is assigned a weight $> 0.8$ by focal loss.

As expected, the many-shot classes had much more hard instances than few-shot or medium-shot classes for almost entire training. As focal loss gives high weights to the hard instances irrespective of their classes, it results in giving more weights to the many-shot classes. As a result, the final result of focal loss is extremely biased towards the many-shot classes as shown in Fig. 5 and the accuracy gap between many-shot and few-shot is $> 60\%$.

Next, we conduct an analysis on how our proposed solution behaves under similar situation. For the analysis, we use the sigmoidal variant of dynamically calculated $\tau$. As our method focuses on class-level difficulty, we classify an instance as hard only if it belongs to a hard class (weight $> 0.8$) and as shown in Fig. 4, we track the average number of hard instances for each subset of classes over the training process. We find that as the training progresses, our method learns to give higher weights to the few-shot and medium-shot classes because the model performs poorly for them. As a result, the average number of instances in many-shot classes that get high weights, gradually becomes lesser than that of medium and few-shot classes leading to better balanced results than focal loss as shown in Fig. 5. For medium-shot and few-shot classes, CDB-W-CE outperforms

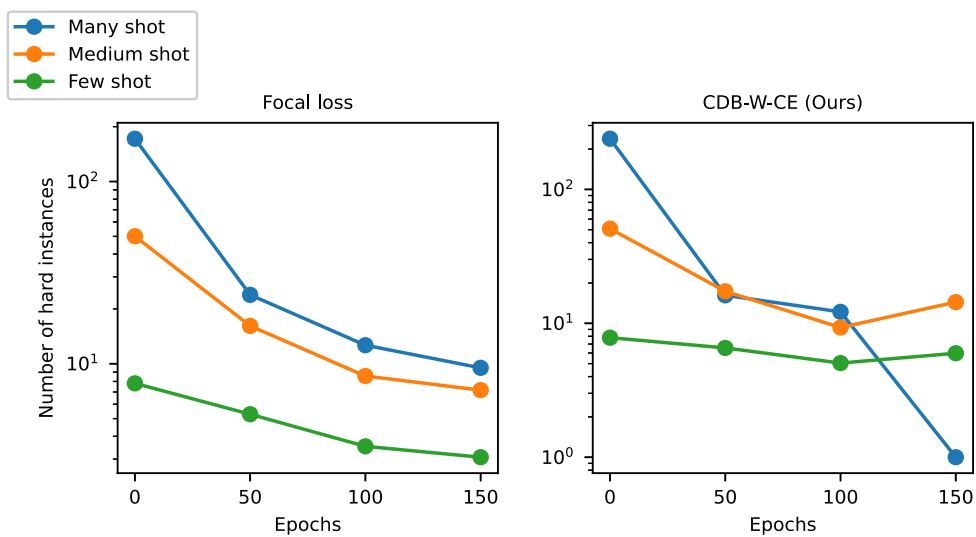

**Fig. 4** Average number of hard instances in many-shot, medium-shot and few-shot classes at various stages of the training for Focal loss and CDB-W-CE (Ours)

focal loss (Lin et al., 2017) by 5.11% and 7.26% in classification accuracy respectively. This helps CDB-W-CE to achieve an overall gain in accuracy of 1.74%.

Note that we do not claim that class-level weighting is always better than instance-level weighting. Instance-level weighting is expected to perform better for datasets with low imbalance and high intra-class variability. For example, if a class 'car' in a dataset has many training samples for blue cars but very few samples for red and green ones, then the model is likely to get biased to blue cars. In such case, instance-level weighting will give high weights to the red and green car samples and try to balance the learning. However, class-level weighting will give same weights to all 'car' samples, still causing bias. In the datasets used in our experiments, we think the intra-class variability is not that high, while the imbalance is extreme causing class-level weighting to perform better.

### 4.2 Object Detection and Instance Segmentation

#### 4.2.1 Datasets

**LVIS** To provide evidence of the effectiveness of our proposed methods on various tasks, we also conduct experiments on the object detection and instance segmentation task. For the purpose, we use the Large Vocabulary Instance Segmentation (LVIS v0.5) (Gupta et al., 2019) due to the long-tailed nature of the dataset. The dataset contains 1,230 categories along with bounding box and instance segmentation annotations. For object detection task, we only use the bounding box annotations but for the instance segmentation task, we use the segmentation mask annotations as well . Gupta et al. (Gupta et al., 2019) proposed a further classification of the categories into three groups based on the number of training images they appear in - frequent ($> 100$ images), common (11-100 images) and rare (1-10 images). We use the same 3 classes to report our evaluation results.

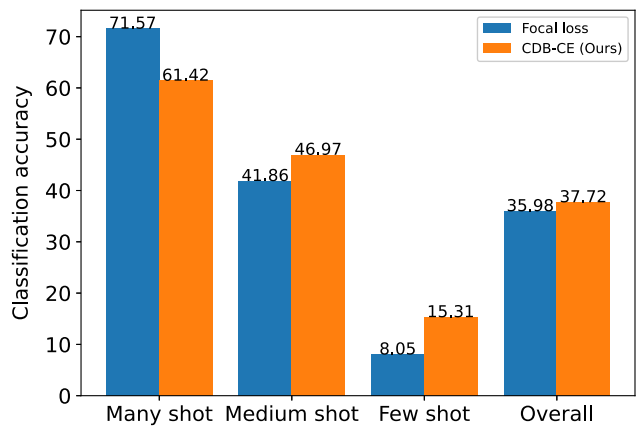

**Fig. 5** Shot-wise accuracy (%) of focal loss v/s CDB-W-CE

#### 4.2.2 Implementation Details

We keep the implementation setup same as (Li et al., 2020). We use Faster R-CNN with ResNet-50-FPN (Ren et al., 2015) for the purpose of object detection. For instance segmentation, we use Mask R-CNN framework with ResNet-50-FPN backbone. The models are trained on four NVIDIA Titan X GPUs for 12 epochs using a SGD optimizer with an initial learning rate of 0.01, momentum of 0.9 and weight decay of 0.0001. The learning rate is decayed by 0.1 after 8 and 11 epochs for both tasks. Gradient clipping and learning rate warm-up are also used.

**Table 5** Object detection results on LVIS v0.5 val set. † means that the results have been copied from the origin paper (Li et al., 2020; Tan et al., 2020)

| Method | Pretrained model | mAP | $AP_r$ | $AP_c$ | $AP_f$ |
| --- | --- | --- | --- | --- | --- |
| CE | trained on MS-COCO | 21.19 | 3.05 | 20.38 | 29.44 |
| CDB-W-CE (Ours) | trained on MS-COCO | 23.37 | 9.83 | 22.89 | 29.53 |
| RFS † (Gupta et al., 2019) | trained on MS-COCO | 23.41 | 14.59 | 22.74 | 27.77 |
| EQL † (Tan et al., 2020) | trained on MS-COCO | 22.80 | 11.30 | 24.70 | 25.10 |
| Focal loss-cls † (Lin et al., 2017) | trained with CE on LVIS | 19.29 | 6.60 | 19.81 | 23.71 |
| LRW-cls † (Li et al., 2020) | trained with CE on LVIS | 24.66 | 14.16 | 23.51 | **30.28** |
| LRW-cls | trained with CDB-W-CE on LVIS | 25.52 | 13.35 | 25.65 | 30.22 |
| BAGS-cls † (Li et al., 2020) | trained with CE on LVIS | 25.96 | 17.65 | 25.75 | 29.54 |
| BAGS-cls | trained with CDB-W-CE on LVIS | 26.01 | 15.47 | **26.52** | 29.58 |
| CDB-W-CE-cls (Ours) | trained with CE on LVIS | 25.40 | 15.54 | 25.50 | 29.21 |
| CDB-W-CE-cls (Ours) | trained with CDB-W-CE on LVIS | **26.31** | **18.15** | 26.37 | 29.52 |

The best results are highlighted in bold

For our method, we use the sigmoidalvariant for dynamic calculation of $\tau$. "-cls"denotes that only the final classification layer is retrained after freezing the other layers using weights from the given pretrained model. To obtain the pretrained model trained with CE or CDB-W-CE on LVIS, we follow exactly the same procedure as Li et al. (2020) i.e., start from a model pretrained on MS-COCO and finetune it on LVIS

**Table 6** Bounding box AP and mask AP ($AP^m$) for image segmentation on LVIS v0.5 val set

| Method | mAP | $AP_r$ | $AP_c$ | $AP_f$ | $mAP^m$ | $AP^m_r$ | $AP^m_c$ | $AP^m_f$ |
| --- | --- | --- | --- | --- | --- | --- | --- | --- |
| Mask R-CNN † | 20.78 | 3.28 | 18.99 | 30.00 | 20.68 | 3.73 | 19.95 | 28.37 |
| Mask-RFS † (Gupta et al., 2019) | – | – | – | – | 24.40 | 14.50 | 24.30 | 28.40 |
| Mask-Calib † (Wang et al., 2019) | – | – | – | – | 21.10 | – | – | – |
| *SimCal* † (Wang et al., 2020) | 22.60 | 13.70 | 20.60 | 28.70 | 23.40 | 16.40 | 22.50 | 27.20 |
| EQL † (Tan et al., 2020) | 23.30 | 11.30 | 24.70 | 25.10 | – | – | – | – |
| BAGS † (Li et al., 2020) | 25.76 | 15.03 | 25.45 | 30.42 | 26.25 | 17.97 | 26.91 | 28.74 |
| Ours | **26.12** | **16.97** | **25.55** | **30.49** | **26.74** | **20.22** | **27.21** | **28.75** |

The best results are highlighted in bold

† denotes that the results have beencopied from the origin paper (Li et al., 2020; Gupta et al., 2019; Wang et al., 2019; Tan et al., 2020; Wang et al., 2020)

### 4.2.3 Results

**Object Detection** For comparison with prior methods, we use 4 evaluation metrics namely mAP (mean average precision over all the classes), $AP_r$ (average precision for rare classes), $AP_c$ (average precision for common classes) and $AP_f$ (average precision for frequent classes). We compare our CDB method with (1) Cross-entropy loss (CE), (2) Repeat factor sampling (RFS) (Gupta et al., 2019), (3) Equalization loss (EQL) (Tan et al., 2020), (4) Focal loss (Lin et al., 2017), (5) Class-wise loss re-weighting (LRW) (Li et al., 2020) and (6) Balanced Group Softmax (BAGS) (Li et al., 2020). We train the detector using different methods and report the results in Table 5.

From Table 5, it can be seen that our CDB-W-CE method outperforms the SOTA methods in mAP and $AP_r$. Evidently, the average precision for the rare classes is significantly boosted by 0.5% using CDB-W-CE. Other than that, we find that our CDB-W-CE method helps to provide better pre-trained models than the normal cross-entropy loss. Using our pretrained model for weight initialization gives a boost in the performance of the SOTA methods such as BAGS and LRW, which is majorly accounted for by a significant improvement for the common classes ($AP_c$). However, in that case we find a drop in performance for the rare classes. But as there are much more common classes than rare classes, the overall performance (mAP) improves.

**Instance Segmentation:** For this task, we compared the performance of CDB-W-CE against that of SOTA methods such as (1) Repeat Factor Sampling (Mask-RFS) (Gupta et al., 2019), (2) Classification Calibration (Mask-Calib) (Wang et al., 2019), (3) *SimCal* (Wang et al., 2020), (4) Equalization loss (EQL) (Tan et al., 2020) (5) Balanced Group Softmax (BAGS) (Li et al., 2020). The results are tabulated in Table 6.

From Table 6, we can see that CDB-W-CE outperforms all previous SOTA methods. Even though CDB-W-CE shows an improvement in all of the metrics, the most significant improvements in performance can be seen for rare classes. The bounding box and mask AP for the rare classes are improved by 1.94% and 2.25% respectively. For the common classes, the improvements are comparatively smaller (0.10% for $AP_c$ and 0.30% for $AP_c^m$). However, there are at least 3 times more common classes than rare ones, resulting the average improvement for common classes to seem small even though the total improvement might be similar to the rare classes. For the frequent classes, even though the improvement by CDB-W-CE is much smaller (0.07% for $AP_f$ and 0.01% for $AP_f^m$), one important thing to note is that we do not sacrifice the performance of the frequent classes to gain for the rare and common classes.

### 4.3 Video-Action Classification

#### 4.3.1 Datasets

**EGTEA** EGTEA (Lin et al., 2018) is an egocentric video dataset that contains a number of trimmed video segments containing multiple kitchen-related actions. The video-action classification task aims to classify each of the video segments to the corresponding action occurring in it. The action labels are combinations of verbs (e.g., put, cut etc.) and nouns (e.g., plate, tomato etc.). Therefore, the action classification task can be further divided into verb classification and noun classification. As the noun classification is very similar to the image classification task, we limit our focus here on the verb classification. EGTEA has 19 different verb classes. The dataset is highly class-imbalanced with an average of 1216 training samples for the 5 most frequent classes and only 158.5 for the rest of the classes. The data distribution is shown in Fig. 6. We conduct experiments on *split1* of the dataset, that comprises of 8299 training video segments and 2022 testing video segments. We separate 1927 video clips from the training data to create our validation data.

#### 4.3.2 Implementation Details

For EGTEA, we train a 3D-ResNeXt101 (Hara et al., 2018) for 100 epochs using batch size of 32 on 8 NVIDIA Titan X GPU's. We use SGD optimizer with momentum 0.9, weight decay 0.0005 and an initial learning rate of 0.001 which is decayed by 0.1 after 60 epochs. For training and testing, we sample 10 RGB frames from each video segment. However, for training, the sampling is done by first dividing the segment into 10 equal sub-segments and then randomly selecting one frame from each of those sub-segments. During testing, the sampling is done at uniform intervals. We use various data augmentation during training, such as random crop, random rotation and horizontal-flipping. For CDB methods, the class-wise difficulty is computed using the validation set. The test set is used only for evaluation.

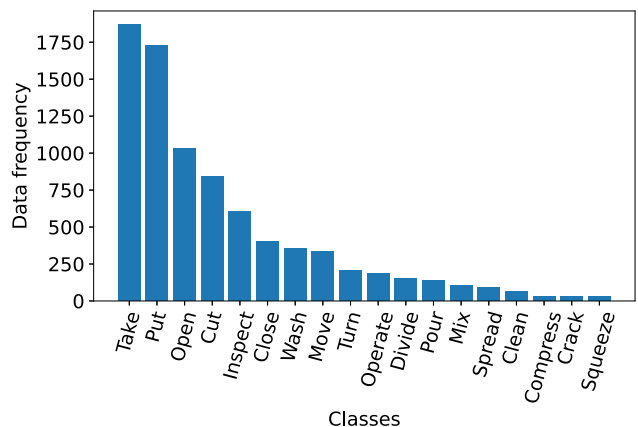

**Fig. 6** Data distribution of EGTEA verbs. The dataset is highly class-imbalanced. For our experiments, we consider the 5 most frequent classes as 'majority classes' and the rest as 'minority classes'

#### 4.3.3 Results on EGTEA

**Comparison to SOTA** We train 3D-ResNeXt on EGTEA verb dataset using different methods e.g., (1) Cross-entropy (CE) loss, (2) Focal loss (Lin et al., 2017), (3) Class-balanced (CB) loss (Cui et al., 2019), (4) CDB-W-CE (Ours) and (5) CDB-S (Ours). We evaluate each of the trained models on the test set and tabulate the results in Table 7. For comparison, we use 4 different metrics. 'Top-1 Acc' and 'Top-5 Acc' are the micro-averaged top-1 and top-5 accuracies for all the 19 classes. 'Precision' and 'Recall' measure the macro-averaged precision and recall for the classes. As the test data is not balanced, the macro-averaged precision and recall values help to verify if the model is biased to certain classes.

As seen from Table 7, our CDB-W-CE and CDB-S methods gives better and balanced performance compared to other SOTA. To analyze further, we compare the performances on the 'head classes' (5 most frequent classes) and the 'tail classes' separately. This helps us to verify that our proposed method causes improvement in the tail classes. The results are tabulated in Table 8. Table 8 verifies that our proposed CDB-W-CE and CDB-S provide an improvement in performance for the tail classes, leading to an overall improvement in performance.

## 5 Conclusion

This paper focuses on the use of class-wise difficulty scores for long-tailed recognition problems. In (Sinha et al., 2020), we proposed a novel way to measure class-wise difficulty and designed a weighted loss function (CDB-W-CE) that determines the class weights from the class difficulty scores. In this paper, we extend on the work and show that class-wise difficulty-based weighting is not limited to only weighted

**Table 7** Results of 3D-ResNeXt101 trained on EGTEA using different methods

|  | Top-1 Acc (%) | Top-5 Acc (%) | Precision (%) | Recall (%) |
| --- | --- | --- | --- | --- |
| CE loss | 67.41 | 95.40 | 61.73 | 64.77 |
| Focal loss (Lin et al., 2017) | 67.46 | 95.66 | 61.94 | 65.12 |
| CB loss (Cui et al., 2019) | 66.86 | 95.69 | 63.39 | 63.26 |
| Freq. based sampling Mikolov et al. (2013) | 66.47 | 96.44 | 63.40 | 66.44 |
| CDB-S (Ours) | 67.85 | 94.51 | 63.59 | **66.56** |
| CDB-W-CE (Ours) | **68.30** | **97.03** | **68.23** | 66.41 |

The best results are highlighted in bold
For CDB-W-CE and CDB-S, we used sigmoidal variant of dynamic calculation for $\tau$

**Table 8** Results of trained 3D-ResNeXt101 on 'head classes' and 'tail classes'

|  | Head classes | | Tail classes | |
| --- | --- | --- | --- | --- |
|  | Precision (%) | Recall (%) | Precision (%) | Recall (%) |
| CE loss | 75.62 | 74.91 | 56.75 | 61.14 |
| Focal loss (Lin et al., 2017) | 75.00 | 70.27 | 53.40 | 55.21 |
| CB loss (Cui et al., 2019) | 73.75 | **75.95** | 59.68 | 58.72 |
| Freq. based sampling (Mikolov et al., 2013) | 74.69 | 75.66 | 59.37 | 63.14 |
| CDB-S (Ours) | 73.33 | 75.18 | 60.11 | **63.48** |
| CDB-W-CE (Ours) | **78.11** | 75.01 | **64.69** | 63.34 |

The best results are highlighted in bold
The five most frequent verb classes (i.e., 'Take','Put', 'Open', 'Cut' and 'Read') are called the 'head classes', while the remaining 14 classes are called the 'tail classes' loss methods. In that direction, we proposed a novel sampling strategy called CDB-S, where samples from each class is sampled with a probability proportional to the class difficulty. Further, given the dependence of class-wise weighting on $\tau$, we proposed 4 different variants for dynamically updating $\tau$ and empirically found that the sigmoidal variant is the reasonable choice irrespective of the dataset or imbalance ratio. We conduct experiments on multiple tasks such as image classification, object detection, instance segmentation and video- action classification. The results showed the consistent effectiveness of CDB-S and CDB-W methods in a wide variety of computer vision tasks.

**Acknowledgements** Computational resource of AI Bridging Cloud Infrastructure (ABCI) provided by National Institute of Advanced Industrial Science and Technology (AIST) was used.

**Author Contributions** All authors contributed to the study conception and design. Data collection, experiments and analysis were performed by SS. The first draft of the manuscript was written by SS and all authors commented on previous versions of the manuscript. All authors read and approved the final manuscript.

**Funding** Not applicable.

**Data Availability** Public Code availability: https://github.com/hitachi-rd-cv/CDB-loss

## Declarations

**Conflict of interest** The authors declare that they have no conflict of interest.

**Ethical Approval** Not applicable.

**Consent to Participate** Not applicable.

**Consent for Publication** Not applicable.